\documentclass[conference]{IEEEtran}
\IEEEoverridecommandlockouts
\usepackage{cite}
\usepackage{amsmath,amssymb,amsfonts}
\usepackage{algorithmic}
\usepackage{graphicx}
\usepackage{textcomp}
\usepackage{tabularx}
\usepackage{xcolor}
\def\BibTeX{{\rm B\kern-.05em{\sc i\kern-.025em b}\kern-.08em
    T\kern-.1667em\lower.7ex\hbox{E}\kern-.125emX}}
\begin{document}

\title{Divorce Prediction with Machine Learning: Insights and LIME Interpretability}

\author{\IEEEauthorblockN{Md Manjurul Ahsan}
\IEEEauthorblockA{\textit{School of Industrial and Systems Engineering} \\
\textit{University of Oklahoma}\\
Norman, OK 73019, USA \\
ahsan@ou.edu}

}

\maketitle

\begin{abstract}
Divorce is one of the most common social issues in developed countries like in the United States. Almost 50\% of the recent marriages turn into an involuntary divorce or separation. While it is evident that people vary to a different extent, and even over time, an incident like Divorce does not interrupt the individual's daily activities; still, Divorce has a severe effect on the individual's mental health, and personal life. Within the scope of this research, the divorce prediction was carried out by evaluating a dataset named by the 'divorce predictor dataset' to correctly classify between married and Divorce people using six different machine learning algorithms- Logistic Regression (LR), Linear Discriminant Analysis (LDA), K-Nearest Neighbors (KNN), Classification and Regression Trees (CART), Gaussian Naïve Bayes (NB), and, Support Vector Machines (SVM).
Preliminary computational results show that algorithms such as SVM, KNN, and LDA, can perform that task with an accuracy of 98.57\%. This work's additional novel contribution is the detailed and comprehensive explanation of prediction probabilities using Local Interpretable Model-Agnostic Explanations (LIME). Utilizing LIME to analyze test results illustrates the possibility of differentiating between divorced and married couples. Finally, we have developed a divorce predictor app considering ten most important features that potentially affect couples in making decisions in their divorce, such tools can be used by any one in order to identify their relationship condition.

\end{abstract}

\begin{IEEEkeywords}
Machine Learning, Statistical Analysis, SVM, KNN
\end{IEEEkeywords}

\section{Introduction}
Divorce also is known as the dissolution of marriage, which is the process of terminating a marriage or marital union \cite{divorce}. While it seems  as simple as  a separation between two people, a divorce may have a significant effect on an individual. A study conducted by the Institute for Population Research \cite{parentdivorce} showed that around 40\% of American children usually experience parental divorce/separation during their childhood and has a long-term effect on their mental health.  Parallel research identifies that children from divorced families received less financial support compared to other children during their lifetime. Additionally, those children also felt more distress due to a lack of emotional support from the families \cite{braver2003relocation}. As a result, parental divorce influences child behavior in a negative manner, which leads to anger, frustration, and depression. 
A divorce could significantly hamper couples, personal and social life. Duncan \& Hoffman (1985) and Morgan (1989)   \cite{duncan1985reconsideration,durcan1989evidence}, showed that  individuals' separation also confronted a variety of stress, financial crisis, reduction in social networks and moving on (Amato, 2000;McLanahan \& Sandefur, 1994)\cite{amato2000consequences,mclanahan1994growing}. Augustine (2000) found that men were nearly 4.8 times as likely to commit suicide as women. They have used the NLMS data for their study. Their study demonstrated a potential relationship between marital status with suicidal risk and,according to the Centers for Disease Control and Prevention (2017), On average, 1.4 million American attempted suicide. The combined medical and work loss are observed as 69 billion dollars \cite{parentdivorce}.  Apart from this, the relationship between divorce and the individuals driving performance such as accident rate, speeding, failure to yield etc. are also highly correlated \cite{mcmurray1970emotional}. In a nutshell, divorce has a significant effect, which brings damage to various aspects such as social, economic, physical, mental and so on.  
During the past few decades, the effect of divorce and the reason behind the divorce/marriage, have been studied thoroughly, while very few studies focused on forecasting the divorce based on some specific criteria \cite{irfan2018comparison,wolfinger2018alternative,yontem2019divorce,li2018markov}. 

Irfan et al. (2018), uses Naïve Bayes and K-Nearest Neighbor methods to predict whether the couple are more likely to get divorce or not. The study took into account several attributes including the plaintiff’s age, age of the defendant, marriage age, child age and so forth. Using those attributes, their computational results show that Naïve Bayes and KNN algorithms can differentiate between divorce and married couples with an accuracy of 72.5\% and 57.5\% respectively. However, the study negates the importance of feature selection procedure, which is one of the most significant drawbacks of this research. In addition, the lower accuracy of the model constitutes the probability of a less accurate model as a decision making tool. 
Another study conducted by Wolfinger (2018) uses a parametric sickle model in order to divorce prediction \cite{wolfinger2018alternative}.

However, none of the referenced literature \cite{irfan2018comparison,wolfinger2018alternative} mentioned regarding the use of standard scale of divorce prediction, that potentially applied in the medical and psychology field. Mustafa et al. (2019) have used correlation-based feature selection and ANN networks in their research. They used the Divorce Predictor Scale (DPS), which is based on Gottman couples' therapy, which includes two class people: divorce (49\%) and married (51\%). Their analysis demonstrated 98.23\% accuracy with Rbf. One of the biggest drawbacks of this research is that the dataset is highly balanced, which ultimately results in the possibility of higher accuracy with high overfitting risk \cite{yontem2019divorce}.

Li (2018) uses the Markov logic Networks Based Method to predict Judicial Decision cases \cite{li2018markov}. However, the study ignored divorce prediction combined with psychological situations, which is thoroughly studied by some other studies \cite{yontem2019divorce}. 

Ibrahim (2020) uses artificial neural network (ANN) methods to predict whether a couple is going to get divorce or not. The model went through multiple learning-validation cycles until it achieved an accuracy of 100\%. Jue et al. (2020) uses three different machine learning algorithms-- support vector machines (svm), random forest (rf), and natural gradient boosting (NGBoost)—on divorce prediction dataset. And their preliminary computational result indicates 98.33\% accuracy on predicting whether the marriage is reliable or not.  Ranjitha et al. (2020) uses particle swarm optimization (PSO) in order to reduce the redundant features that do not contribute to the divorce prediction. However, none of the study results explain in details of feature selection or discuss why their proposed model should be trusted.
 
 While data mining is often employed by the majority of the study in psychology and psychiatry, one cannot conclude that it is adequate for a phenomenon with long-term consequences, such as divorce. Fortunately, if divorce predictors can be used in advance, many of the divorces would be avoided.
Taking these opportunities into account, firstly, this paper analyzed six different machine learning algorithms that evaluate whether a couple might get divorce or not, using “divorce predictor dataset” Secondly, top 10 features out of 54 features were identified that highly correlate with the divorce prediction. Thirdly, the prediction probabilities are analyzed using Local Interpretable Model-Agnostic Explanations (LIME). Finally, an desktop based graphical user interface (GUI) application is developed using tkinter and machine learning model that can be accessed and used by individual to determine whether he/she is more likely get divorce in the future, ultimately presenting the opportunity to take necessary steps in saving conjugal life.

\section{Methodology}
\begin{figure}[h]
\centerline{\includegraphics[width=1.5 in]{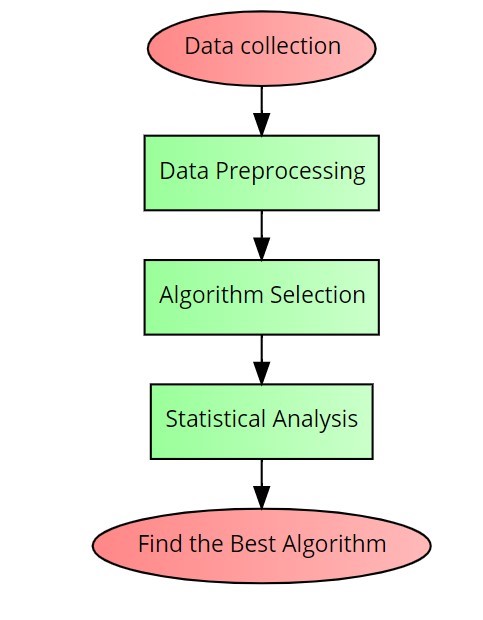}}
\caption{Experiment Methodology.}
\label{fig:fig1}
\end{figure}

Figure~\ref{fig:fig1} represents the overall research methodology used in this study.

\subsection{Data collection and preprocessing}
We adapted the dataset named by “Divorce predictor dataset” from UCI machine learning Repository which is one of the most popular open sources for dataset \cite{yontem2019divorce}. Table \ref{table:tab1} summarizes the data set characteristics in details, that contains 170 instances and 54 attributes. All attributes have been added to index A. 

\renewcommand{\tabcolsep}{1pt}
\begin{table}[h]
\centering
\caption{Divorce Predictor Dataset Characteristics }
\label{table:tab1}
\begin{tabular}{|l|l|} 
\hline
\textbf{Dataset Characteristics}       & Multivariate, Univariate  \\ 
\hline
\textbf{Attribute~Characteristics}     & Integer                   \\ 
\hline
\textbf{Associated Tasks}              & Classification            \\ 
\hline
\textbf{Area}                          & Life                      \\ 
\hline
\textbf{Number of Web Hits}            & 40288                     \\ 
\hline
\textbf{\textbf{Number of Instances}}  & 170                       \\ 
\hline
\textbf{\textbf{Number of attributes}} & 54                        \\ 
\hline
\textbf{\textbf{Missing Values}}       & N/A                       \\ 
\hline
\textbf{\textbf{Date Donated}}         & 2019-07-24                \\
\hline
\end{tabular}
\end{table}

\begin{figure}[h]
\centerline{\includegraphics[width=3.5 in]{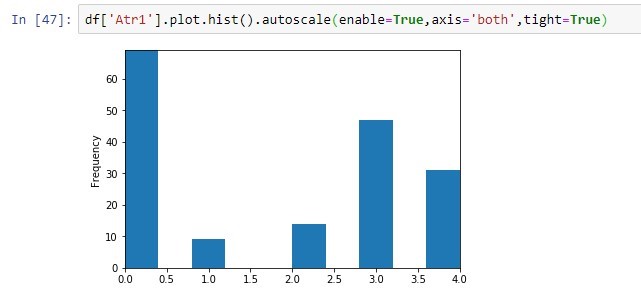}}
\caption{A sample example of Histogram of first attribute with 170 instances.}
\label{fig:fig2}
\end{figure}

Figure~\ref{fig:fig2} shows a histogram of first attribute with 170 instances. All 54 histograms have been added to Appendix A. 

We dedicated 80\% of the data for the training and remaining 20\% for the testing purpose. Figure 2 presents a set of representative data used in this study.

\begin{figure}[h]
\centerline{\includegraphics[width=3.8 in]{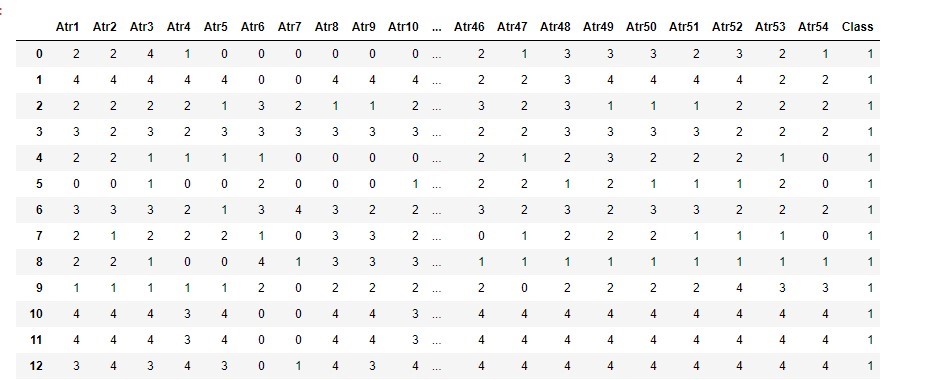}}
\caption{Sample Example Divorce Predictor Dataset.}
\label{fig:fig3}
\end{figure}

We applied six widely used machine learning algorithms: Logistic Regression(LR), Linear Discriminant Analysis(LDA), K-Nearest Neighbors(KNN), Classification and Regression Trees(CART), Gaussian Naive Bayes(NB) and Support Vector Machines(SVM) on “Divorce Predictor Dataset”.  The experiment was carried out using an office grade laptop with the specification as follows: Intel Core I7, 16 GB RAM. We have used 10 fold cross validation and the computational results are presented as the average score of all folds.

\subsection{Results}
Table~\ref{tab:taaab} presents a summary of the performance of all algorithms on ``Divorce predictor dataset”. SVM outperformed all algorithms in terms of higher accuracy and lower error rate. In contrast, CART showed the worst performance across all measures.

\renewcommand{\tabcolsep}{1pt}
\label{tab:taaab}
\begin{table}[h]
\centering
\caption{Divorce Predictor Dataset Characteristics }
\begin{tabular}{|l|l|l|} 
\hline
\textbf{Algorithm} & \textbf{Accuracy} & \textbf{Error}  \\ 
\hline
LR~ ~ ~            & 0.978             & 0.33            \\ 
\hline
LDA~ ~ ~           & 0.985             & 0.028           \\ 
\hline
KNN                & 0.985             & 0.028           \\ 
\hline
CART               & 0.963~            & 0.048           \\ 
\hline
NB                 & 0.978             & 0.04            \\ 
\hline
SVM                & 0.99              & 0.02            \\
\hline
\end{tabular}
\end{table}

We further explore the SVM algorithms performance with newly developed dataset by computing accuracy, precision, recall and f-1 score, which presents in Table 3.

\renewcommand{\tabcolsep}{1pt}
\begin{table}[h]
\centering
\caption{Divorce Predictor Dataset Characteristics }
\begin{tabular}{|l|l|l|l|l|} 
\hline
             & \textbf{Precision} & \textbf{Recall} & \textbf{f1-score} & \textbf{Support}  \\ 
\hline
\textbf{0}   & 0.90               & 1.00            & 0.95              & 19                \\ 
\hline
\textbf{1}   & 1.00               & 0.87            & 0.93              & 15                \\ 
\hline
\textbf{Avg} & 0.95               & 0.95            & 0.94              & 34(Total)         \\
\hline
\end{tabular}
\end{table}

\subsection{Explainable AI}
\begin{figure}[h]
\centerline{\includegraphics[width=3.5 in]{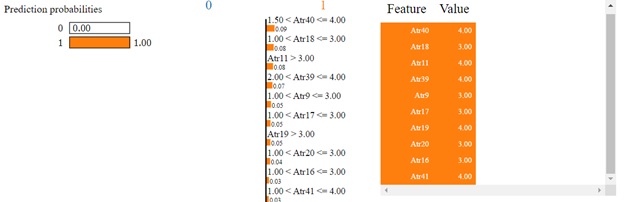}}
\caption{Divorce predictor application.}
\label{fig:fig41}
\end{figure}

Local interpretable model agnostic (LIME) being used to identify important features that helps SVM model to make decision during the prediction. Figure~\ref{fig:fig41} presents top 10 features that helps SVM algorithm to correctly identify the married couples. Using sklearn tool such as selectkbest methods we have reevaluated the top 10 features, which are similar as found using explainable LIME.

Finally, a graphical user interface (GUI) based divorce predictor app is developed using top 10 important features from Gottman couples therapy which can be accessed by individual who are more likely identify their pre-existing relationship status.

\begin{figure}[h]
\centerline{\includegraphics[width=3.5 in]{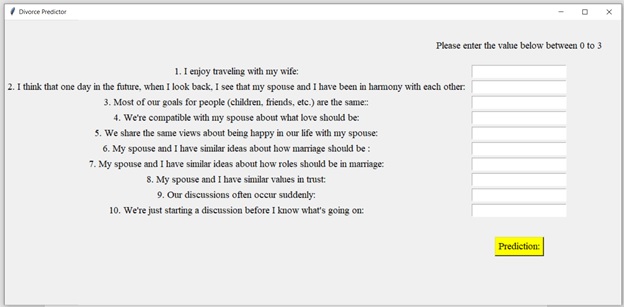}}
\centerline{\includegraphics[width=3.5 in]{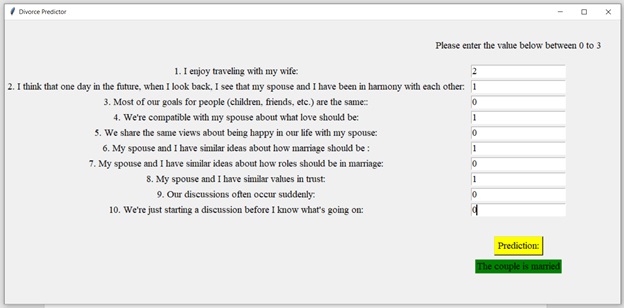}}
\caption{Gottman couple therapy based app for divorce prediction using Tkinter GUI.}
\label{fig:fig4}
\end{figure}

\section{Discussion and potential remarks}
When the overall results of the study are examined, it was observed that the Divorce Predictors Scale developed within the scope of Gottman couples therapy can predict divorce rates with an accuracy of around 99\% using SVM. Gottman and Silver (2015) stated that the decisive factor in satisfaction of couples with sexuality, love and passion in their marriages depends on the quality of friendship between husband and wife by 70\%. In addition, factors such as value systems of couples, their views of love, perceptions and attitudes about trust are important for creating a shared meaning~cite{gottman2014predicts}. 

We investigated a dataset which contains 54 attributes with two different classes (Divorced, married couples) with 170 instances. . We have used 6 different algorithms in order to predict the probability of whether the couple is more likely getting divorce or not. Our study shows that it is possible to achieve 99\% accuracy with SVM. Using scikit-learn selectkbest methods and LIME, top 10 features are identified. Finally, a GUI based application is developed for individuals using the python Tkinter tool. Researcher, practitioner and psychiatrist can use such tools to evaluate the patient's current state regarding divorce and can take necessary steps to prevent the divorce.
However, the dataset used in this study is highly balanced, ultimately resulting in higher accuracy during the preliminary computations.  In future the current research will focus on applying other machine learning and deep learning approaches to evaluate the study results more in details, and experiment with high dimensional data.

\bibliographystyle{IEEEtran}
\bibliography{refs}
\begin{appendices}
\section{1. Attribute}
\begin{enumerate}
  \item When one of our apologies apologizes when our discussions go in a bad direction, the issue does not extend.
  \item I know we can ignore our differences, even if things get hard sometimes.
  \item When we need it, we can take our discussions with my wife from the beginning and correct it.
  \item When I argue with my wife, it will eventually work for me to contact him. 
  \item The time I spent with my wife is special for us. 
  \item We don’t have time at home as partners.
  \item We are like two strangers who share the same environment at home rather than family.
  \item I enjoy our holidays with my wife. 
  \item I enjoy traveling with my wife.
  \item My wife and most of our goals are common. 
  \item I think that one day in the future, when I look back, I see that my wife and I are in harmony with each other. 
  \item My wife and I have similar values in terms of personal freedom.
  \item My husband and I have similar entertainment.
  \item Most of our goals for people (children, friends, etc.) are the same. 
  \item Our dreams of living with my wife are similar and harmonious.
  \item We’re compatible with my wife about what love should be.
  \item We share the same views with my wife about being happy in your life .
  \item My wife and I have similar ideas about how marriage should be.
  \item My wife and I have similar ideas about how roles should be in marriage.
  \item My wife and I have similar values in trust 
  \item I know exactly what my wife likes.
  \item I know how my wife wants to be taken care of when she’s sick. \item I know my wife’s favorite food.
  \item I can tell you what kind of stress my wife is facing in her life.
  \item I have knowledge of my wife’s inner world.
  \item I know my wife’s basic concerns. 
  \item I know what my wife’s current sources of stress are.
  \item I know my wife’s hopes and wishes.
  \item I know my wife very well.
  \item I know my wife’s friends and their social relationships.
  \item I feel aggressive when I argue with my wife. 
  \item When discussing with my wife, I usually use expressions such as are you always or are you never. 
  \item I can use negative statements about my wife’s personality during our discussions.
  \item I can use offensive expressions during our discussions.
  \item I can insult our discussions.
  \item I can be humiliating when we argue.
  \item My argument with my wife is not calm.
  \item I hate my wife’s way of bringing it up. 
  \item Fights often occur suddenly.
  \item We’re just starting a fight before I know what’s going on. 
  \item When I talk to my wife about something, my calm suddenly breaks.
  \item When I argue with my wife, it only snaps in and I don’t say a word.
  \item I’m mostly thirsty to calm the environment a little bit.
  \item Sometimes I think it’s good for me to leave home for a while.
  \item I’d rather stay silent than argue with my wife.
  \item Even if I’m right in the argument, I’m thirsty not to upset the other side. 
  \item When I argue with my wife, I remain silent because I am afraid of not being able to control my anger. 
  \item I feel right in our discussions. 
  \item I have nothing to do with what I’ve been accused of.
  \item I’m not actually the one who’s guilty about what I’m accused of.
  \item I’m not the one who’s wrong about problems at home.
  \item I wouldn’t hesitate to tell her about my wife’s inadequacy. 
  \item When I discuss it, I remind her of my wife’s inadequate issues.
 
  \item I’m not afraid to tell her about my wife’s incompetence.
\end{enumerate}
The above dataset can be downloaded 
from http://archive.ics.uci.edu/ml/datasets/Divorce+Predictors+data+set 
\end{appendices}
\end{document}